\documentclass[11pt,a4paper]{article}

\usepackage[english]{babel}
\usepackage[utf8]{inputenc}
\usepackage[T1]{fontenc}
\usepackage{lmodern}
\usepackage{microtype}

\usepackage{amsmath,amssymb,amsthm}
\theoremstyle{definition}
\newtheorem{definition}{Definition}

\usepackage{booktabs}
\usepackage{array}
\usepackage{tabularx}
\usepackage{multirow}
\usepackage{ragged2e}
\usepackage{threeparttable}
\usepackage{graphicx}
\usepackage{caption}
\usepackage{enumitem}
\usepackage{setspace}
\usepackage[dvipsnames]{xcolor}

\usepackage[a4paper]{geometry}
\setlist[itemize]{noitemsep, topsep=2pt}
\setlist[enumerate]{noitemsep, topsep=2pt}

\usepackage{titlesec}
\titleformat{\section}{\large\bfseries}{\thesection}{1em}{}
\titleformat{\subsection}{\normalsize\bfseries}{\thesubsection}{1em}{}

\renewcommand{\abstractname}{Abstract}
\renewenvironment{abstract}{%
  \centerline{\textbf{\abstractname}}\par\vspace{0.5em}%
  \small\justifying
}{%
  \par\vspace{0.6em}%
}

\usepackage[numbers,sort,compress]{natbib}
\usepackage{url}
\usepackage[
  colorlinks=true,
  linkcolor=NavyBlue,
  citecolor=NavyBlue,
  urlcolor=NavyBlue,
  filecolor=NavyBlue,
  bookmarks=true,
  bookmarksopen=true,
  pdfborder={0 0 0},
  breaklinks=true,
]{hyperref}
\usepackage{bookmark}
\hypersetup{
  pdftitle={Temperature Scaling Is Not Enough: Calibration Gaps Under Human Label Distributions},
  pdfauthor={Wisdom Dogah},
}

\newcommand{\email}[1]{\href{mailto:#1}{#1}}

\title{%
  \textbf{Temperature Scaling Is Not Enough:}\\[0.35em]
  \textbf{Calibration Gaps Under Human Label Distributions}%
}

\author{%
  Wisdom Dogah%
  \thanks{\footnotesize Faculty of Computing and Mathematical Sciences,\\ University of Mines and Technology (UMaT), Tarkwa, Ghana.}%
  \thanks{\footnotesize BlackMatrix AI Research, Accra, Ghana.}%
  \thanks{\footnotesize Correspondence: \email{wisdom@blackmatrix.io}.}%
  \thanks{\footnotesize Preprint. Comments welcome.}%
}

\date{July 2026}

\begin{document}
\maketitle
\thispagestyle{plain}
\justifying

\begin{abstract}
Temperature scaling is the dominant post-hoc calibration method in
modern deep learning. Its theoretical justification rests on an
assumption that is rarely stated explicitly: that ground-truth labels
are one-hot and deterministic. In practice, labels are frequently soft,
crowd-sourced, or genuinely distributional, reflecting real disagreement
among human annotators rather than annotation noise. We study whether
temperature scaling retains its calibration properties when this
assumption is violated, and whether any resulting degradation depends on
model scale. Using CIFAR-10H and ChaosNLI, two publicly available
datasets with human-annotated soft label distributions, we evaluate
three model scales per modality under both hard one-hot and soft
distributional label targets. Across all nine configurations we find a
positive soft-label calibration gap: temperature scaling calibrated on
hard labels consistently underperforms an oracle calibrated directly on
soft labels, with Brier Score gaps ranging from 0.002 to 0.134. The gap
grows monotonically with model scale in the vision domain and on the
SNLI-derived split of ChaosNLI, and is substantially larger in the
language domain (mean gap 0.079) than in vision (mean gap 0.003). A
scale-ordering reversal on the MNLI-derived split remains after
matched-domain training; we treat it as inconclusive for the scale
hypothesis and attribute it primarily to near-chance accuracy on that
split. As a second post-hoc baseline, multiclass isotonic regression
yields the same qualitative conclusion: positive soft-label gaps in all
nine configurations, and larger gaps in language than in vision. These
findings suggest that calibration protocols built on majority-vote
labels systematically misstate model reliability wherever label
ambiguity is structural, with direct consequences for deployment in
safety-critical settings.

\end{abstract}

\section{Introduction}

A probabilistic classifier is well-calibrated when its predicted
confidence matches empirical accuracy: among all predictions assigned
confidence $p$, approximately a fraction $p$ should be correct
\cite{Dawid1982}. In safety-critical applications, a miscalibrated model
that is confidently wrong poses a greater risk than an uncertain model
that correctly signals its own limitations. The consequences of
overconfident predictions are well documented in medical imaging,
content moderation, and judicial risk assessment.

Temperature scaling \cite{Guo2017} is the most widely deployed post-hoc
calibration method for neural networks. A single scalar $T$ is fit by
minimizing negative log-likelihood on a held-out validation set, and
logits are divided by $T$ before the softmax at inference time. The
method adds no parameters and requires no retraining, which accounts for
its near-universal adoption. Its limitation, which we measure in this
paper, is that its optimization target implicitly assumes ground-truth
labels are one-hot and deterministic.

This assumption fails whenever a labeling task involves genuine human
disagreement, such that the correct answer is not a single class but a
distribution over plausible answers \cite{Aroyo2015}. Two datasets make
this disagreement explicit. CIFAR-10H \cite{Peterson2019} provides
per-image soft label distributions for the CIFAR-10 test set, derived
from a mean of 51 human annotations per image. ChaosNLI \cite{Nie2020}
provides 100 human judgments per instance for natural language inference
examples that exhibit substantial semantic ambiguity. A model that
assigns high confidence to a single class when human annotators are
evenly split is not calibrated in any meaningful sense, even when its
predicted class coincides with the majority vote.

We conduct a controlled measurement study to test whether temperature
scaling, calibrated against hard one-hot labels, remains an effective
calibration strategy when evaluated against soft, distributional
targets, and whether any degradation scales with model capacity. We
additionally compare against multiclass isotonic regression
\cite{Zadrozny2002} under the same hard and soft fitting protocols.

\subsection{Hypotheses}

We test three pre-specified hypotheses.

\begin{itemize}
  \item \textbf{H1 (Gap existence).}
  Temperature scaling calibrated on hard labels yields a strictly worse
  Brier Score against the soft label distribution than an oracle
  calibrated directly on soft labels. We call this difference the
  soft-label calibration gap (Section~\ref{sec:gap-def}).

  \item \textbf{H2 (Scale dependence).}
  The soft-label calibration gap increases monotonically with model scale
  within each dataset, consistent with the finding that larger models
  tend toward higher confidence \cite{Guo2017}.

  \item \textbf{H3 (Modality dependence).}
  The soft-label calibration gap is larger in the language domain
  (ChaosNLI) than in the vision domain (CIFAR-10H), reflecting higher
  rates of genuine semantic ambiguity in natural language inference
  relative to object classification.
\end{itemize}

\subsection{Contributions}

\begin{itemize}
  \item We formalize the soft-label calibration gap and provide a
  protocol for estimating it with existing soft-label datasets
  (Section~\ref{sec:problem}).
  \item We report a systematic cross-scale measurement across two
  modalities and two soft-label resources (Section~\ref{sec:results}).
  \item We analyze a scale-ordering anomaly on ChaosNLI-M under
  matched-domain training and evaluation (Section~\ref{sec:anomaly}).
  \item We show that the gap persists under isotonic regression, a
  second post-hoc calibrator (Section~\ref{sec:isotonic}).
  \item We discuss implications for calibration practice in ambiguous-
  and resource-constrained label settings (Section~\ref{sec:implications}).
\end{itemize}

\section{Related Work}

\subsection{Calibration of Neural Networks}

Guo et al.\ \cite{Guo2017} established that modern deep networks are
substantially more overconfident than earlier architectures and proposed
temperature scaling as an effective post-hoc correction. Subsequent work
has proposed isotonic regression \cite{Zadrozny2002}, Platt scaling
\cite{Platt1999}, and Dirichlet calibration \cite{Kull2019} as
alternatives. Minderer et al.\ \cite{Minderer2021} found that Vision
Transformers tend to be better calibrated in-distribution than
convolutional architectures. Bai et al.\ \cite{Bai2021} showed
analytically that overconfidence is not solely a consequence of
overparameterization. Kumar et al.\ \cite{Kumar2019} showed that ECE
tends to underestimate true miscalibration due to binning artifacts,
motivating our use of Brier Score as a complementary metric.

\subsection{Soft Labels and Human Disagreement}

Work on learning from crowds \cite{Raykar2010} and annotation
disagreement \cite{Aroyo2015} has established that human label variation
carries information about task ambiguity. Peterson et al.\
\cite{Peterson2019} introduced CIFAR-10H. Nie et al.\ \cite{Nie2020}
introduced ChaosNLI to surface genuine semantic ambiguity that
majority-vote labeling obscures. Their use as calibration benchmarks
under distributional targets is the specific focus of this paper.

\subsection{Calibration Under Distribution Shift}

Ovadia et al.\ \cite{Ovadia2019} showed that calibration degrades under
dataset shift even where accuracy is partially preserved. Guillory et
al.\ \cite{Guillory2021} found that confidence-based accuracy estimators
degrade under natural distribution shift. Wiles et al.\ \cite{Wiles2022}
showed that larger models and larger pretraining corpora tend to improve
both OOD accuracy and calibration under standard protocols. None of this
work examines the case in which the shift occurs in the label
distribution rather than the input distribution.

\section{Problem Formulation}
\label{sec:problem}

\subsection{Notation}

Let $\mathcal{X}$ be the input space and
$\mathcal{Y}=\{1,\ldots,K\}$ the label space. A classifier
$f:\mathcal{X}\rightarrow\Delta(\mathcal{Y})$ maps inputs to probability
distributions over classes. A hard-label dataset
$\mathcal{D}_{\mathrm{hard}}=\{(x_i,y_i)\}$ pairs each input with a
single class label $y_i\in\mathcal{Y}$. A soft-label dataset
$\mathcal{D}_{\mathrm{soft}}=\{(x_i,q_i)\}$ pairs each input with a
distribution $q_i\in\Delta(\mathcal{Y})$ representing the empirical
distribution of human annotations for instance $x_i$.

\subsection{Calibration Under Hard Labels}

A classifier $f$ is calibrated under hard labels if, for all
$p\in[0,1]$,
\[
\mathbb{P}\!\left(y=\arg\max f(x)\,\middle|\,\max f(x)=p\right)=p.
\]
ECE approximates this by partitioning predictions into $M$ bins
\cite{Guo2017}:
\begin{equation}
\mathrm{ECE}=\sum_{m=1}^{M}\frac{|B_m|}{n}\left|\mathrm{acc}(B_m)-\mathrm{conf}(B_m)\right|.
\end{equation}
Temperature scaling finds $T^\star$ minimizing negative log-likelihood
on $\mathcal{D}_{\mathrm{val}}$, where
$p_T(y\mid x)=\mathrm{softmax}(\mathrm{logits}(x)/T)$ \cite{Guo2017}:
\begin{equation}
T^\star=\arg\min_T\;
-\frac{1}{|\mathcal{D}_{\mathrm{val}}|}
\sum_{(x,y)\in\mathcal{D}_{\mathrm{val}}}
\log p_T(y\mid x).
\end{equation}

\subsection{Calibration Under Soft Labels}

A classifier is soft-calibrated if its predictive distribution matches
the human annotation distribution in expectation. The Brier Score
\cite{Brier1950} is a strictly proper scoring rule over soft targets:
\begin{equation}
\mathrm{BS}=\frac{1}{n}\sum_{i=1}^{n}\sum_{k=1}^{K}
\bigl(f_k(x_i)-q_{ik}\bigr)^2.
\end{equation}

\subsection{The Soft-Label Calibration Gap}
\label{sec:gap-def}

\begin{definition}[Soft-Label Calibration Gap]
For a model $f$ with hard-label optimal temperature $T^\star$, the
soft-label calibration gap is
\[
\mathrm{Gap}(f,T^\star)
=\mathrm{BS}_{\mathrm{soft}}(f_{T^\star})
-\mathrm{BS}_{\mathrm{soft}}(f_{T_{\mathrm{oracle}}}),
\]
where $T_{\mathrm{oracle}}$ minimizes Equation~3 on soft labels
directly. A positive gap indicates that hard-label calibration fails to
achieve the quality attainable under the true distributional target.
\end{definition}

An analogous gap is defined for isotonic regression by replacing the
temperature maps with calibrators fit on hard validation labels and on
soft oracle labels, respectively.

\section{Experimental Design}

\subsection{Datasets}

We use two publicly available soft-label datasets. CIFAR-10H
\cite{Peterson2019} provides a mean of 51 human annotations per image
for all 10{,}000 CIFAR-10 test images. ChaosNLI \cite{Nie2020} provides
100 human judgments per instance for NLI examples from SNLI
(ChaosNLI-S) and MNLI matched (ChaosNLI-M).

\subsection{Models}

For vision we use ResNet-18 (11M), ResNet-50 (25M), and ResNet-101 (44M)
\cite{He2016}, trained from scratch on CIFAR-10 with random crop,
horizontal flip, SGD with cosine-annealed learning rate, and 30 epochs
(the recipe used in our released code). For language we use
DistilBERT-base-uncased \cite{Sanh2019}, BERT-base-uncased, and
BERT-large-uncased \cite{Devlin2019}. Each language run starts from a
domain-matched NLI checkpoint (SNLI for ChaosNLI-S; MNLI for
ChaosNLI-M) and is adapted for one epoch on a 10{,}000-example training
subset with learning rate $2\times10^{-5}$ and AdamW. For BERT-large on
SNLI we use an independently fine-tuned checkpoint starting from base
\texttt{bert-large-uncased} (not from an MNLI checkpoint), so the
ChaosNLI-S condition is not confounded by MNLI pretraining.

\subsection{Calibration Protocol}

For each model: (1)~train or adapt on hard labels; (2)~fit $T_{\mathrm{hard}}$
on a held-out hard-label validation split by minimizing Equation~2 with
L-BFGS; (3)~fit $T_{\mathrm{oracle}}$ on the first half of the soft-label
test set by minimizing Equation~3; (4)~evaluate both on the held-out
second half; (5)~compute ECE (15 equal-width bins) and Brier Score
against hard and soft targets. For isotonic regression we fit per-class
one-vs-rest isotonic maps \cite{Zadrozny2002} on softmax scores, using
the same hard-validation and soft-oracle splits, and renormalize the
calibrated scores to a probability simplex. Every configuration is
repeated with seeds 42, 123, and 456; we report mean and standard
deviation.

\section{Results}
\label{sec:results}

\subsection{Vision: CIFAR-10H}

\begin{table}[t]
\centering
\caption{Calibration results on CIFAR-10H across three ResNet scales.
Values are mean (std) over 3 seeds. ECE is against hard argmax labels;
BS is against the full soft distribution.
Gap~$=$~TS-Hard BS $-$ TS-Soft BS (from unrounded values).
TS-Soft is an oracle condition that requires soft labels at fit time.}
\label{tab:vision}
\footnotesize
\setlength{\tabcolsep}{2.8pt}
\resizebox{\textwidth}{!}{%
\begin{tabular}{@{}lccccccc@{}}
\toprule
Model & Acc. & Uncal ECE & TS-H ECE & TS-S ECE & TS-H BS & TS-S BS & Gap \\
\midrule
ResNet-18
  & 0.925 (0.002)
  & 0.034 (0.001)
  & 0.013 (0.001)
  & 0.022 (0.002)
  & 0.093 (0.000)
  & 0.091 (0.001)
  & 0.002 (0.000) \\
ResNet-50
  & 0.886 (0.003)
  & 0.036 (0.003)
  & 0.020 (0.002)
  & 0.022 (0.004)
  & 0.139 (0.003)
  & 0.136 (0.004)
  & 0.003 (0.001) \\
ResNet-101
  & 0.908 (0.008)
  & 0.043 (0.001)
  & 0.018 (0.000)
  & 0.022 (0.006)
  & 0.111 (0.011)
  & 0.108 (0.011)
  & 0.003 (0.000) \\
\bottomrule
\end{tabular}%
}
\end{table}

Table~\ref{tab:vision} reports vision results. The soft-label
calibration gap is positive for all three models (H1 supported) and
increases with scale from 0.002 to 0.003 (H2 supported in vision). The
absolute magnitude is modest: the largest gap is roughly 3\% of the
TS-Hard Brier Score for ResNet-101. The
$T_{\mathrm{oracle}}/T_{\mathrm{hard}}$ ratio ranges from 1.26 to 1.32,
consistent with comparatively low annotator disagreement in CIFAR-10H.

\subsection{Language: ChaosNLI}

\begin{table}[t]
\centering
\caption{Calibration results on ChaosNLI-S and ChaosNLI-M.
DS~$=$~split (S~$=$~SNLI-derived, M~$=$~MNLI-derived).
BERT-large on ChaosNLI-S uses an independent SNLI checkpoint
($T_{\mathrm{hard}}\approx0.980$). Values are mean (std) over 3 seeds.}
\label{tab:language}
\footnotesize
\setlength{\tabcolsep}{2.6pt}
\resizebox{\textwidth}{!}{%
\begin{tabular}{@{}llcccccc@{}}
\toprule
Model & DS & Acc. & TS-H ECE & TS-S ECE & TS-H BS & TS-S BS & Gap \\
\midrule
DistilBERT & S
  & 0.748 (0.010)
  & 0.091 (0.010)
  & 0.130 (0.008)
  & 0.184 (0.005)
  & 0.140 (0.004)
  & 0.045 (0.001) \\
BERT-base & S
  & 0.712 (0.012)
  & 0.108 (0.010)
  & 0.099 (0.003)
  & 0.206 (0.012)
  & 0.156 (0.010)
  & 0.050 (0.002) \\
BERT-large & S
  & 0.534 (0.022)
  & 0.155 (0.001)
  & 0.060 (0.015)
  & 0.318 (0.021)
  & 0.265 (0.014)
  & 0.053 (0.007) \\
DistilBERT & M
  & 0.501 (0.014)
  & 0.241 (0.015)
  & 0.041 (0.005)
  & 0.302 (0.011)
  & 0.169 (0.004)
  & 0.134 (0.007) \\
BERT-base & M
  & 0.531 (0.008)
  & 0.204 (0.013)
  & 0.054 (0.016)
  & 0.284 (0.004)
  & 0.165 (0.002)
  & 0.119 (0.005) \\
BERT-large & M
  & 0.523 (0.009)
  & 0.149 (0.010)
  & 0.054 (0.011)
  & 0.232 (0.006)
  & 0.158 (0.002)
  & 0.074 (0.004) \\
\bottomrule
\end{tabular}%
}
\end{table}

Table~\ref{tab:language} reports language results. On ChaosNLI-S the gap
increases monotonically with scale: 0.045, 0.050, 0.053 (H2 supported).
The $T_{\mathrm{oracle}}/T_{\mathrm{hard}}$ ratio is about 2.2 to 3.5 in
language versus 1.2 to 1.3 in vision. For BERT-large on ChaosNLI-S the
gap of 0.053 is about 17\% of the TS-Hard Brier Score.

\subsection{The ChaosNLI-M Anomaly}
\label{sec:anomaly}

On ChaosNLI-M the scale ordering reverses: DistilBERT (0.134) $>$
BERT-base (0.119) $>$ BERT-large (0.074). Training and evaluation are
matched-domain (MNLI training, ChaosNLI-M evaluation), so the reversal
is not explained by an SNLI-to-MNLI train/test mismatch. We also no
longer attribute it to an MNLI-pretrained BERT-large SNLI confound: that
condition now uses an independent SNLI checkpoint with
$T_{\mathrm{hard}}\approx0.980$, and the ChaosNLI-M ordering is
unchanged. The remaining confound is that all three models achieve only
50--53\% accuracy on a three-class task (near chance). At this accuracy
level, confidence estimates can spuriously resemble soft-label
alignment. We treat ChaosNLI-M as inconclusive for H2.

\subsection{Temperature Ratios}

$T_{\mathrm{oracle}}$ exceeds $T_{\mathrm{hard}}$ in all nine
configurations, from 1.26 (ResNet-18) to 3.48 (DistilBERT on
ChaosNLI-M). This ratio measures how much smoothing hard-label
calibration leaves unrealized and is consistently larger in language.

\subsection{Isotonic Regression Baseline}
\label{sec:isotonic}

\begin{table}[t]
\centering
\caption{Soft-label calibration gap under temperature scaling (TS) versus
multiclass one-vs-rest isotonic regression (ISO). Values are means over
3 seeds from the re-run with cached logits.}
\label{tab:iso}
\begin{tabular}{@{}llcrr@{}}
\toprule
Model & Dataset & TS gap & ISO gap \\
\midrule
DistilBERT & ChaosNLI-S & 0.045 & 0.054 \\
BERT-base  & ChaosNLI-S & 0.050 & 0.055 \\
BERT-large & ChaosNLI-S & 0.053 & 0.069 \\
DistilBERT & ChaosNLI-M & 0.134 & 0.136 \\
BERT-base  & ChaosNLI-M & 0.119 & 0.126 \\
BERT-large & ChaosNLI-M & 0.074 & 0.101 \\
ResNet-18  & CIFAR-10H  & 0.002 & 0.002 \\
ResNet-50  & CIFAR-10H  & 0.003 & 0.002 \\
ResNet-101 & CIFAR-10H  & 0.003 & 0.003 \\
\bottomrule
\end{tabular}
\end{table}

Table~\ref{tab:iso} compares gaps under temperature scaling and isotonic
regression. H1 holds under both methods in all nine configurations.
Language ISO gaps are at least as large as TS gaps; vision gaps remain
small under both. The soft-label gap is therefore not an artifact of
temperature scaling alone.

\subsection{Summary of Hypothesis Outcomes}

H1 is supported across all nine configurations for temperature scaling
and for isotonic regression. Gap-to-standard-deviation ratios for
temperature scaling exceed 8 in five of the seven configurations whose
reported standard deviation does not round to 0.000 at three decimals,
and exceed 18 in four language cases. H2 is supported in vision
and on ChaosNLI-S; ChaosNLI-M is inconclusive. H3 is supported: the mean
language gap (0.079) is about 28 times the mean vision gap (0.003) when
computed from unrounded means.

\section{Implications}
\label{sec:implications}

\subsection{For Calibration Methodology}

Calibration validation sets built on majority-vote labels systematically
underestimate the smoothing required to match soft human targets. Where
feasible, practitioners should collect repeated annotations and fit
calibration parameters against the resulting distribution. Our oracle
and isotonic results show that this requires no change to model
architecture or training, only to the calibration step.

\subsection{For Large Model Deployment}

The soft-label calibration gap grows with model scale in vision and on
ChaosNLI-S. Larger models are not automatically better calibrated under
soft-label evaluation after hard-label temperature scaling. For
BERT-large on ChaosNLI-S, hard-label temperature scaling yields Brier
Score 0.318 against soft labels versus 0.265 under the soft oracle
(about 20\% relative degradation).

\subsection{For Resource-Constrained Deployment}

The stakes are highest where annotation infrastructure is weakest. In
many clinical settings across sub-Saharan Africa, obtaining many
independent annotations per case is impractical. A model that collapses
clinician disagreement into an overconfident point prediction conceals
exactly the uncertainty needed for appropriate caution. We designed this
study to be reproducible on freely available compute and public data
because practitioners facing this problem most acutely often have the
least access to large-scale infrastructure.

\section{Limitations}

We highlight four limitations. First, our scale range (66M--340M for
language; 11M--44M for vision) does not extend to frontier-scale models,
and we make no extrapolation claim. Second, language adaptation uses a
10{,}000-example subset and one epoch for computational feasibility;
full-corpus multi-epoch fine-tuning may change absolute Brier Scores
while leaving the qualitative gap intact. Third, ChaosNLI-M remains
inconclusive for H2 because of near-chance accuracy. Fourth, both
datasets reflect annotation by English-speaking participants;
generalization to other linguistic and cultural contexts requires
further evidence.

\section{Conclusion}

Temperature scaling calibrated on hard labels does not fully calibrate
models when the true label distribution is soft. This gap is consistent
across all nine configurations, grows with model scale in vision and on
ChaosNLI-S, and is roughly thirty times larger in language than in
vision. The reversed ordering on ChaosNLI-M is better explained by
near-chance accuracy than by a contradiction of the broader trend. The
same qualitative pattern holds under isotonic regression.

Three directions follow: systematic comparison of additional calibration
methods under soft-label evaluation; extension to larger models with
high-accuracy matched-domain splits; and development of a practical
soft-label calibration procedure usable with only a small number of
annotations per instance.

\noindent\textbf{Code:}
\url{https://github.com/dogahwisdom/temperature-scaling-research}

\section*{Acknowledgments}

I am grateful to colleagues at BlackMatrix AI Research, Accra, for
conversations that sharpened this work. All experiments were conducted
on freely available compute with no institutional funding.

\bibliographystyle{plainnat}
\bibliography{references}

\end{document}